\documentclass[runningheads]{llncs}
\usepackage[T1]{fontenc}
\usepackage{graphicx,verbatim}
\usepackage{multirow}
\usepackage{bbding}
\usepackage{amssymb}
\usepackage{amsmath}
\usepackage[normalem]{ulem}
\useunder{\uline}{\ul}{}
\usepackage{booktabs}
\usepackage{bm}
\usepackage{url}
\usepackage[table,xcdraw]{xcolor}
\usepackage{xcolor}
\usepackage{threeparttable}
\usepackage{hyperref}
\hypersetup{
    colorlinks=true, 
    urlcolor=magenta,   
}

\usepackage[misc]{ifsym}

\usepackage[marginal]{footmisc}

\newcommand{\figref}[1]{Fig.~\ref{#1}}
\newcommand{\tabref}[1]{Table~\ref{#1}}
\newcommand{\eqtref}[1]{Eq.~\ref{#1}}

\begin{document}
\title{Prototype-Guided Cross-Modal Knowledge Enhancement for Adaptive Survival Prediction} 
\titlerunning{ProSurv for Adaptive Survival Analysis}

\author{Fengchun Liu\inst{1} \and 
Linghan Cai\inst{1} \and 
Zhikang Wang\inst{2} \and 
Zhiyuan Fan\inst{1} \and 
Jin-gang Yu\inst{3}\textsuperscript{(\Letter)} \and 
Hao Chen\inst{4}\textsuperscript{(\Letter)} \and 
Yongbing Zhang\inst{1}\textsuperscript{(\Letter)}}

\authorrunning{F. Liu et al.}

\institute{
\text Harbin Institute of Technology(Shenzhen), Shenzhen, China\\ 
\email{ybzhang08@hit.edu.cn}
\and
Monash University, Melbourne, Australia
\and
South China University of Technology, Guangzhou, China\\
\email{jingangyu@scut.edu.cn}
\and
The Hong Kong University of Science and Technology, Hong Kong, China\\
\email{jhc@cse.ust.hk}}
\maketitle
\footnote{F. Liu, L. Cai, and Z. Wang contributed equally to this work.}
\begin{abstract}
Histo-genomic multimodal survival prediction has garnered growing attention for its remarkable model performance and potential contributions to precision medicine. However, a significant challenge in clinical practice arises when only unimodal data is available, limiting the usability of these advanced multimodal methods. To address this issue, this study proposes a prototype-guided cross-modal knowledge enhancement (ProSurv) framework, which eliminates the dependency on paired data and enables robust learning and adaptive survival prediction. Specifically, we first introduce an intra-modal updating mechanism to construct modality-specific prototype banks that encapsulate the statistics of the whole training set and preserve the modality-specific risk-relevant features/prototypes across intervals. Subsequently, the proposed cross-modal translation module utilizes the learned prototypes to enhance knowledge representation for multimodal inputs and generate features for missing modalities, ensuring robust and adaptive survival prediction across diverse scenarios. Extensive experiments on four public datasets demonstrate the superiority of ProSurv over state-of-the-art methods using either unimodal or multimodal input, and the ablation study underscores its feasibility for broad applicability. Overall, this study addresses a critical practical challenge in computational pathology, offering substantial significance and potential impact in the field. Codes are available at \href{https://github.com/cyclexfy/ProSurv}{https://github.com/cyclexfy/ProSurv}.



\end{abstract}

\section{Introduction}
Survival analysis seeks to estimate the risk of specific events, e.g. death or disease recurrence; consequently, it is crucial for disease progression estimation and treatment strategy selection in clinical practice \cite{jiang2024autosurv,xiang2025vision}. Recently, whole-slide pathological images (WSIs) and genomic data have been widely used to model patient characteristics, with the first and second types of data capturing microscopic morphology features and quantitative molecular information, respectively. 

Existing computational pathology methods can be roughly categorized as: WSI-based unimodal, genome-based unimodal, and histo-genomic multimodal methods. The unimodal methods, either WSI-based \cite{ilse2018attention,lu2021data,shao2021transmil,wang2023surformer,wang2024dual} or genome-based \cite{klambauer2017self} ones, have demonstrated significant success in model performance and interpretation. However, the intrinsic complexity and heterogeneity of tumors urge the use of multimodal data to provide a more comprehensive characterization of patients. Born out of necessity, numerous multimodal approaches \cite{chen2021multimodal,jaume2024modeling,xu2023multimodal,zhou2023cross} have been developed to leverage the complementary and shared information provided by different data modalities. These approaches achieve better prediction accuracy by effectively combining the strengths of both modalities.

Although multimodal approaches have an advantage over unimodal ones in performance, they face a significant challenge in clinical practice \cite{wang2025histo}, i.e., these developed methods do not apply to patients without complete data. This scenario is particularly common in current clinical systems, where patients often transfer between hospitals during the diagnostic phase or may not undergo specific examinations due to suboptimal conditions \cite{ning2022mutual}. 
From the methodology perspective, there are two potential solutions. First, using knowledge distillation \cite{chen2021learning,hu2020knowledge} to transfer multimodal knowledge to unimodal networks. However, this is constrained by low training efficiency and flexibility. Second, direct cross-modal reconstruction for multimodal knowledge completion is infeasible due to inherent heterogeneity between pathology and genomic modality \cite{li2022hfbsurv}. 

To address the challenges, this paper proposes a prototype-guided cross-modal knowledge enhancement method (ProSurv). Specifically, through an intra-modal updating mechanism, we construct modality-specific prototype banks that capture risk-relevant features across intervals and encapsulate statistics of the whole training set.
Afterward, the proposed cross-modal translation module utilizes prototypes as guidance to enhance knowledge representation for multimodal inputs and generate missing modality features for unimodal input. Consequently, our method eliminates the dependency on paired data and enables robust learning and adaptive survival prediction in diverse scenarios.
We extensively evaluate ProSurv on four public datasets and demonstrate the superiority of our method over state-of-the-art methods using either unimodal or multimodal input. The ablation study underscores its feasibility for broad applicability.

\section{Methodology}
\begin{figure}[!t]
\centerline{\includegraphics[width=0.95\columnwidth]
{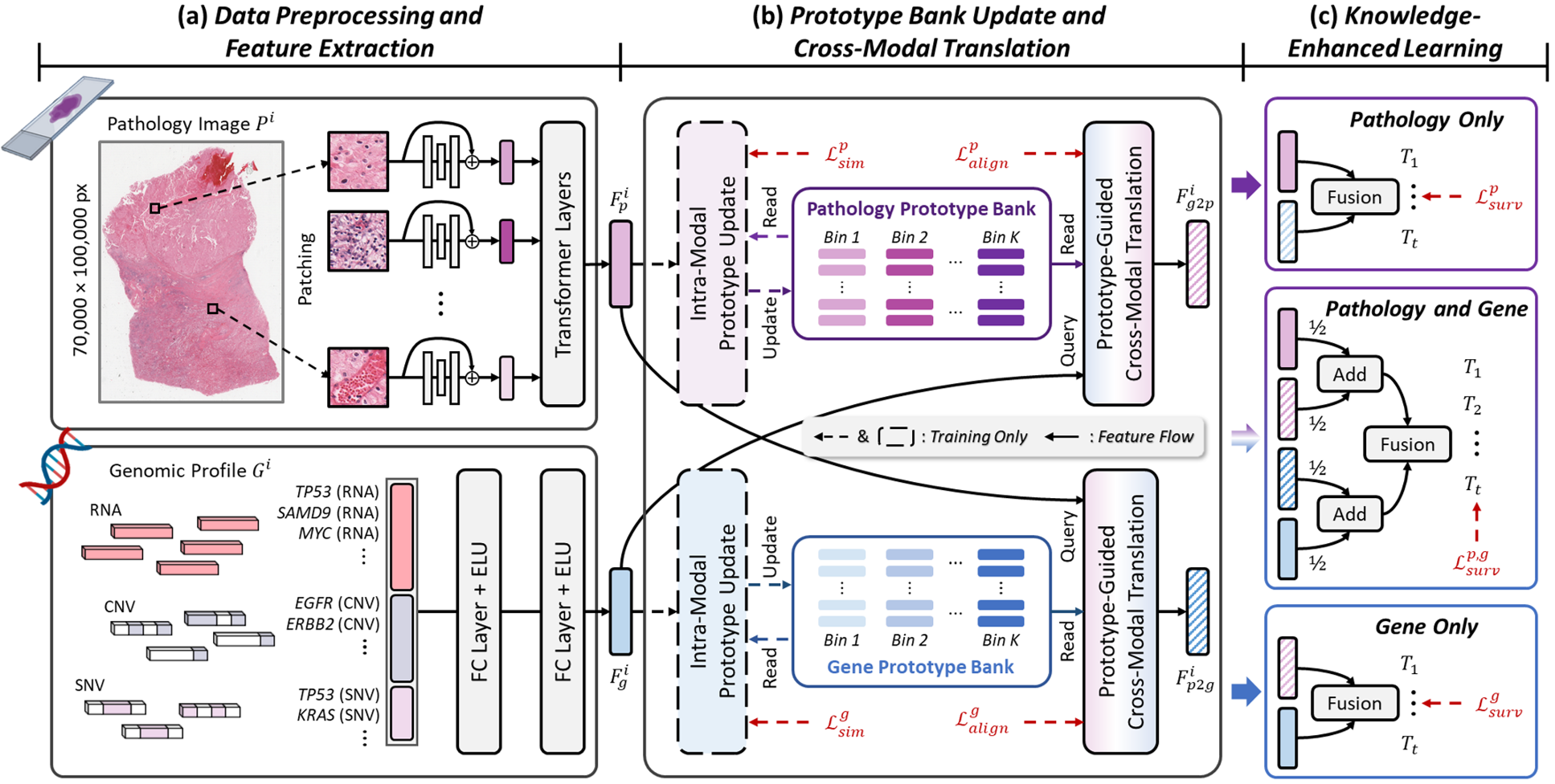}}
\caption{Overview of the ProSurv. We first proceed with data preprocessing and feature extraction for WSIs and genomic data (a). Afterward, prototype banks are established corresponding to the input modalities and prototype-guided cross-modal translation modules enable the cross-modal feature translation (b). Eventually, the features from input data and translated modules jointly achieve the adaptive survival analysis (c). }
\label{fig:overview}
\vspace{-0.25cm}
\end{figure}

\figref{fig:overview} illustrates the framework of ProSurv, which comprises three steps including data preprocessing and feature extraction, prototype bank update and cross-modal translation, and knowledge-enhanced learning. The following subsections introduce them in detail. 


\subsection{Problem Formulation}
Survival analysis \cite{clark2003survival} aims to predict time-to-event outcomes, where the event may not be fully observed. For the $i$-th patient, we model the survival and hazard functions as $f^{i}_{\text{surv}}(T\geq t\mid X^i)$ and $f^{i}_{\text{hazard}}(T=t\mid T\geq t, X^i)$, respectively. Here, $X^i=\{P^i, G^i, c^i, t^i\}$ represents the patient's data, where $P^i$, $G^i$, and $c^i \in \{0,1\}$ are the WSI, genomic profiles, and censorship status, respectively, and $t^i \in \mathbb{R}^+$ indicates overall survival (in months). Here, survival time $t^i$ is discretized into $K$ intervals $\{bin_1,...,bin_K\}$, allowing the model to estimate the hazard function for each bin. The survival function is approximated as: $f^{i}_{\text{surv}}(t_k) = \prod_{u=1}^{k}(1-f_{\text{hazard}}^i(t_u))$, where $t_k$ corresponds to the $k$-th bin. We learn the representation $f(X^i)$ to compute the survival loss $\mathcal{L}_{\text{surv}}(f(X^i), t^i, c^i)$ using the negative log-likelihood (NLL) loss function \cite{zadeh2020bias}. 
\vspace{-0.1cm}

\subsection{Data Preprocessing and Feature Extraction}
\noindent \textbf{Pathological Image Feature.} 
Given an input WSI $P^i$ from the $i$-th patient, we remove the non-tissue regions and then crop the rest into non-overlapping patches. The preprocessed data can be denoted as $\{p^i_j\}_{j=1}^{N^i}$, where $N^i$ represents the total patch number. The patches are then fed into a pre-trained feature extractor ${E}_p(\cdot)$ to obtain a bag of patch features. Afterward, $L$ Transformer layers \cite{vaswani2017attention,xiong2021nystromformer} are employed to model relations between patches, followed by a global average pooling (GAP) to derive the WSI-level representation. The whole process can be formulated as: $F^i_p=\text{GAP}(\text{Transformer}^{(L)}(\{{E}_p(p^i_j)\}_{j=1}^{N^i}))\in \mathbb{R}^{1\times d}$.

\noindent \textbf{Genomic Feature.}
The input genomic data $G^i$ consists of numerous $1\times 1$ measurements $\{g^i_j\}_{j=1}^{M}$, where $M$ means gene panel range. A Self-Normalizing Neural Network \cite{klambauer2017self} $E_g(\cdot)$ is applied for feature extraction. This process can be formulated as $F^{i}_{g} = E_g(\{g^i_j\}_{j=1}^{M})\in \mathbb{R}^{1\times d}$, where $d$ is the feature dimension. 
\vspace{-0.1cm}

\begin{figure}[!t]
\centerline{\includegraphics[width=0.95\columnwidth]
{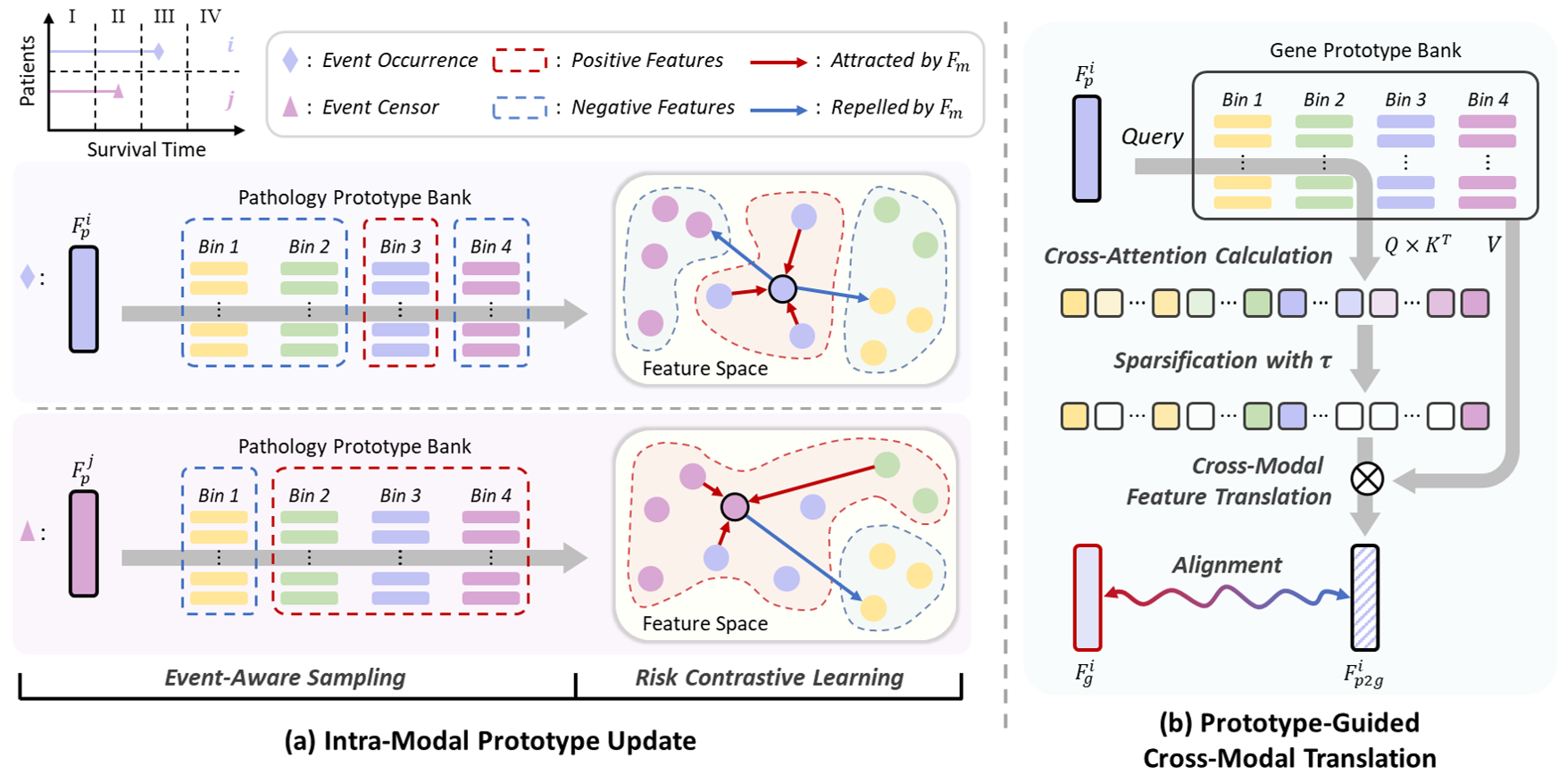}}
\caption{Details of the intra-modal prototype update (a) and prototype-guided cross-modal translation (b). Here, we employ pathological features as an input for demonstration.}
\label{fig:module}
\vspace{-0.25cm}
\end{figure}

\subsection{Intra-Modal Prototype Update}
To preserve modality-specific knowledge, we construct pathological prototype bank $B_p \in \mathbb{R}^{K \times n \times d}$ and genomics prototype bank $B_g \in \mathbb{R}^{K \times n \times d}$ corresponding to the two modalities. Here, $K$ is identical to the number of intervals and $n$ is the number of prototypes in each bank. To enable prototypes with more risk-relevant knowledge, we develop an intra-modal prototype updating mechanism (\figref{fig:module}(a)).

Given the extracted features from the $i$-th patient, we calculate the cosine similarity between the features and the corresponding prototypes recorded in the bank. Taking the pathological feature $F_p^i$ as an example, we can obtain:
\begin{equation}
w_p^{i,k}=\frac{1}{K} \sum_{j=1}^nw_{p}^{i,k,j},\ \text{where}\  w_{p}^{i,k,j} = \frac{\text{Norm}(F^i_p)\cdot \text{Norm}(B^{k, j}_p)}{\|\text{Norm}(F^i_p)\| \|\text{Norm}(B^{k, j}_p)\|}.
\label{eq1}
\end{equation}
In \eqtref{eq1}, $\text{Norm}(\cdot)$ is a min-max normalization, $\| \cdot \|$ denotes the Euclidean Norm, $B_p^{k,j}$ indicates the $j$-th prototype in the $k$-th bin of the pathology prototype bank, $w_p^{i,k}$ measures the average value between the input data and the prototypes of the $k$-th bin. To enhance prototype representations, we employ risk contrastive learning, which pulls prototypes from the same time interval closer to the input feature and pushes those from different intervals apart. Considering the complexity of survival data, we introduce event-aware sampling. Specifically, for uncensored data, prototypes from the bin corresponding to the label are treated as positive samples, while others are negative. For censored data, the censoring time is treated as a key node: prototypes before the censoring time are negative samples, while those after it are positive, as the patient is alive before the censoring time. Hence, the entire procedure can be summarized as follows:
\begin{equation}
\mathcal{L}^p_{\text{sim}}= (1-c^i)(-{w_{p}^{t^i}} + \sum_{t^k \neq t^i} \frac{w_{p}^{t^k}}{K-1}) + c^i( -\sum_{t^k \geq t^i} \frac{w^{t^k}}{K-t^i+1} + \sum_{t^k < t^i} \frac{w^{t^k}}{t^i-1}),
\label{eq2}
\end{equation}
where $t^i$ represents the the bin corresponding to the event occurrence. 

In the training process, the intra-modal prototype update mechanisms are applied to each modality, which can be formulated as: $\mathcal{L}^{p,g}_{\text{sim}}= \mathcal{L}^p_{\text{sim}} + \mathcal{L}^g_{\text{sim}}$.

\subsection{Prototype-Guided Cross-Modal Translation}
The prototype-guided cross-modal translation aims to utilize the input features along with the prototype banks to reconstruct cross-modal features, as shown in \figref{fig:module}(b). The function is mathematically achieved through a cross-attention mechanism \cite{vaswani2017attention}. Here, we take the $i$-th patient as an example. When reconstructing genomic features, pathological features $F^{i}_{p}$ serves as the query $Q_p$ to interact with keys $K_g$, the genomic prototype bank, producing attention scores reflecting histo-genomics associations. Afterward, the values $V_g$ derived from the genomic bank are multiplexed with the attention scores, generating the final reconstructed cross-modal features. The whole process can be formulated as follows: 
\begin{equation}\small
F^i_{p2g} = \sigma\left( \frac{(W^q_p F^i_p)(W^k_g B_g)^T}{\tau \sqrt{d}} \right) (W^v_g B_g), 
\label{eq3}
\end{equation}
where $\sigma(\cdot)$ is a softmax function, $T$ and $\tau$ represent a transpose operation and a temperature coefficient, respectively. Similarly, translating genomic features $F^{i}_{g}$ to the pathological ones can be formulated as:
$
F^i_{g2p} = \sigma\left( \frac{(W^q_g F^i_g)(W^k_p B_p)^T}{\tau \sqrt{d}} \right) (W^v_p B_p). 
$

To enhance the learning process, we selectively align reconstructed features with their original counterparts when the complete modality pairs are available. The corresponding loss $\mathcal{L}_{\text{align}}$ is defined as:
\begin{equation}
\mathcal{L}^{p,g}_{\text{align}}=\mathcal{L}^p_{\text{align}} +  \mathcal{L}^g_{\text{align}}=\|F^i_p-F^i_{g2p}\|^2 + \|F^i_g-F^i_{p2g}\|^2.
\label{eq5}
\end{equation}

\subsection{Knowledge-Enhanced Learning and Prediction}
Knowledge-enhanced learning (\figref{fig:overview}(c)) is designed to conduct adaptive survival prediction regardless of the input data conditions. Here, we introduce two cases: the patient with complete data modalities and incomplete data modality.

\noindent \textbf{Complete Modalities.} With the complete modalities, we can obtain the original features $F^i_p$, $F^i_g$, and the translated features $F^i_{g2p}$, $F^i_{p2g}$. Then, we fuse the original features with the corresponding translated features using an averaging function and then utilize the concatenated features as the final patient representation for survival prediction. The process can be formulated as follows: 
\begin{equation}
H_{p, g} = \phi(\text{Concat}(F^i_{ep}, F^i_{eg})),\ \text{where} \ F^i_{ep}=\frac{F^i_{p} + F^i_{g2p}}{2},\ F^i_{eg}=\frac{F^i_{g} + F^i_{p2g}}{2}.
\label{eq6}
\end{equation}
Here, $\phi(\cdot)$ is a full-connected (FC) layer followed by a sigmoid function, $\text{Concat}$ means concatenation. Under this scenario, the training loss is expressed as: $\mathcal{L}^{p,g}_{\text{total}} = \mathcal{L}^{p,g}_{\text{surv}} + \alpha \mathcal{L}^{p,g}_{\text{sim}} + \beta \mathcal{L}^{p,g}_{\text{align}}$, where $\alpha$ and $\beta$ are hyperparameters for balance.

\begin{table}[!t]\scriptsize
\centering
\caption{Performance comparison using C-index (mean $\pm$ standard deviation) on four cancer datasets. The best result is shown in {\color{red}{red}}, and the second-best one is in {\color{blue}{blue}}. ``P'' and ``G'' are abbreviations of pathology and gene, respectively. ``$\ast$'' represents the multimodal training and unimodal testing scenario. }
\vspace{-0.15cm}
\resizebox{0.85\linewidth}{!}{%
\begin{tabular}{l|cc|cccc|c}
\toprule
\multirow{2}{*}{Methods} & \multicolumn{2}{c|}{Modal} & \multicolumn{4}{c|}{Datasets} & \multirow{2}{*}{Overall} \\ \cmidrule{2-3} \cmidrule{4-7}
& P & G & BRCA & BLCA & STAD & CRAD &                          \\ \midrule
ABMIL & \checkmark & $\times$ & 0.613$\pm$0.118 & 0.575$\pm$0.050 & 0.561$\pm$0.059 & 0.586$\pm$0.115 & 0.584                          \\
TransMIL& \checkmark & $\times$ & 0.625$\pm$0.080 & 0.614$\pm$0.046 & 0.522$\pm$0.050 & 0.525$\pm$0.118 & 0.572   \\
WIKG & \checkmark & $\times$ & 0.613$\pm$0.086 & 0.585$\pm$0.047 & 0.551$\pm$0.049 & 0.517$\pm$0.127 & 0.569   \\ 
MambaMIL & \checkmark & $\times$ & 0.617$\pm$0.078 & 0.612$\pm$0.058 & 0.574$\pm$0.041 & 0.588$\pm$0.095 & 0.598   \\ 
\midrule
SNN  & $\times$ & \checkmark & 0.576$\pm$0.112 & 0.539$\pm$0.052 & 0.539$\pm$0.032 & 0.616$\pm$0.090 & 0.568   \\
SNNTrans & $\times$ & \checkmark & 0.551$\pm$0.097 & 0.553$\pm$0.043 & 0.555$\pm$0.057 & 0.614$\pm$0.125 & 0.568   \\ \midrule
MCAT & \checkmark & \checkmark & 0.654$\pm$0.077 & 0.616$\pm$0.055 & 0.572$\pm$0.098 & 0.615$\pm$0.141 & 0.614   \\
MOTCAT& \checkmark & \checkmark & 0.665$\pm$0.111 & 0.600$\pm$0.051 & 0.563$\pm$0.084 & 0.593$\pm$0.169 & 0.605  \\
CMTA& \checkmark & \checkmark   & 0.604$\pm$0.045 & 0.642$\pm$0.072 & 0.577$\pm$0.098 & 0.586$\pm$0.075 & 0.602   \\
SurvPath & \checkmark & \checkmark     & 0.675$\pm$0.069 & 0.584$\pm$0.040 & 0.551$\pm$0.040 & 0.585$\pm$0.100 & 0.599   \\ \midrule
G-HANet & \checkmark & $\ast$ & \color{blue}{0.677$\pm$0.071} & 0.603$\pm$0.054 & 0.588$\pm$0.050 & 0.598$\pm$0.169 & 0.617   \\ \midrule
\cellcolor[HTML]{EFEFEF} & \cellcolor[HTML]{EFEFEF}\checkmark & \cellcolor[HTML]{EFEFEF}$\ast$ & \cellcolor[HTML]{EFEFEF}\color{red}{0.701$\pm$0.101} & \cellcolor[HTML]{EFEFEF}0.611$\pm$0.032 & \cellcolor[HTML]{EFEFEF}\color{red}{0.620$\pm$0.069} & \cellcolor[HTML]{EFEFEF}\color{blue}{0.626$\pm$0.090} & \cellcolor[HTML]{EFEFEF}\color{blue}{0.640} \\
\cellcolor[HTML]{EFEFEF}  & \cellcolor[HTML]{EFEFEF}$\ast$ & \cellcolor[HTML]{EFEFEF}\checkmark & \cellcolor[HTML]{EFEFEF}0.578$\pm$0.110 & \cellcolor[HTML]{EFEFEF}\color{blue}{0.646$\pm$0.046} & \cellcolor[HTML]{EFEFEF}0.540$\pm$0.079 & \cellcolor[HTML]{EFEFEF}0.595$\pm$0.056 & \cellcolor[HTML]{EFEFEF}0.590 \\
\multirow{-3}{*}{\cellcolor[HTML]{EFEFEF}ProSurv} \cellcolor[HTML]{EFEFEF}  & \cellcolor[HTML]{EFEFEF}\checkmark & \cellcolor[HTML]{EFEFEF}\checkmark & \cellcolor[HTML]{EFEFEF}0.675$\pm$0.070 & \cellcolor[HTML]{EFEFEF}\color{red}{0.655$\pm$0.047} & \cellcolor[HTML]{EFEFEF}\color{blue}{0.609$\pm$0.073} & \cellcolor[HTML]{EFEFEF}\color{red}{0.641$\pm$0.088} & \cellcolor[HTML]{EFEFEF}\color{red}{0.645} \\ \bottomrule
\end{tabular}}
\label{tab:quantitative}
\vspace{-0.3cm}
\end{table}

\noindent \textbf{Incomplete Modality.} When the patient only has one kind of data, ProSurv also can perform robust learning and prediction. For example, with only WSI input, ProSurv can generate the translated genomic feature $F^i_{p2g}$ from pathological features $F^i_{p}$. Afterward, we fuse these features for the final prediction $H_p$:
\begin{equation}
H_p = \phi(\text{Concat}(F^i_{p}, F^i_{p2g})).
\label{eq6}
\end{equation}
Vice versa as:
$
H_g = \phi(\text{Concat}(F^i_{g}, F^i_{g2p})).
$
The training loss under the incomplete data input is considered: $\mathcal{L}^{p}_{\text{total}} = \mathcal{L}^{p}_{\text{surv}} + \alpha \mathcal{L}^{p}_{\text{sim}}$ (only pathology input) or $\mathcal{L}^{g}_{\text{total}} = \mathcal{L}^{g}_{\text{surv}} + \alpha \mathcal{L}^{g}_{\text{sim}}$ (only gene input).
\section{Experiments and Results}
\subsection{Datasets \&  Evaluation Metric \& Implementation Details}
Experiments were performed on four public datasets from The Cancer Genome Atlas (TCGA). Specifically, we used 868 cases of Breast Invasive Carcinoma (BRCA), 359 cases of Bladder Urothelial Carcinoma (BLCA), 318 cases of Stomach Adenocarcinoma (STAD), and 294 cases of Colon and Rectum Adenocarcinoma (CRAD). All WSIs were processed at 20$\times$ magnification using CLAM \cite{lu2021data} and genomic data was processed using min-max normalization. For each dataset, we randomly split the data into training, validation, and test sets with a ratio of 6:2:2, and reported the average C-index \cite{harrell1982evaluating} across 5-fold cross-validation. 

ProSurv is implemented on PyTorch 1.12.1 \cite{paszke2019pytorch} using an NVIDIA RTX 4090 GPU. Patch-level features are extracted using the UNI model \cite{chen2024uni}. 4,096 patches out of the whole image are randomly selected from each WSI during training, while all patches are used for validation and testing. The Adam optimizer \cite{kingma2014adam} is employed for model optimization with a constant learning rate of 1e-4 and a weight decay of 1e-4. Each model is trained for a maximum of 50 epochs, and the best-performing model on the validation set is used for testing. 
We set the hyper-parameters as: $K=4$, $n=32$, $\tau=0.5$, $\alpha=0.2$, and $\beta=0.2$. 

\begin{table}[!t]\scriptsize 
\centering
\caption{Ablation study. For each testing scenario, the best performance is highlighted in {\color{red}{red}}, while the second-best performance is in {\color{blue}{blue}}. ``w/o'' means without.}
\resizebox{0.90\linewidth}{!}{%
\begin{tabular}{l|cc|cccc|c}
\toprule
\multirow{2}{*}{Methods} & \multicolumn{2}{c|}{Modal} & \multicolumn{4}{c|}{Datasets} & \multirow{2}{*}{Overall} \\ \cmidrule{2-3} \cmidrule{4-7}
& P & G & BRCA & BLCA & STAD & CRAD &                          \\ \midrule
w/o Prototypes & \checkmark & $\ast$ & 0.660$\pm$0.068 & 0.601$\pm$0.041 & 0.503$\pm$0.047 & 0.617$\pm$0.111 & 0.595                          \\
w/o $\mathcal{L}_{\text{sim}}$ & \checkmark & $\ast$ & 0.639$\pm$0.069 & {\color{red}0.618$\pm$0.043} & 0.533$\pm$0.104 & {\color{blue}0.618$\pm$0.131} & 0.602                          \\
w/o $\mathcal{L}_{\text{align}}$ & \checkmark & $\ast$ & {\color{blue}0.687$\pm$0.079} & {\color{blue}0.616$\pm$0.042} & {\color{blue}0.534$\pm$0.100} & 0.616$\pm$0.128 & {\color{blue}{0.613}}                          \\
\cellcolor[HTML]{EFEFEF}ProSurv & \cellcolor[HTML]{EFEFEF}\checkmark & \cellcolor[HTML]{EFEFEF}$\ast$ & \cellcolor[HTML]{EFEFEF}{\color{red}{0.701$\pm$0.101}} & \cellcolor[HTML]{EFEFEF}{0.611$\pm$0.032} & \cellcolor[HTML]{EFEFEF}{\color{red}{0.620$\pm$0.069}} & \cellcolor[HTML]{EFEFEF}{\color{red}{0.626$\pm$0.090}} & \cellcolor[HTML]{EFEFEF}{\color{red}{0.640}}                          \\ \midrule \midrule
w/o Prototypes & $\ast$ & \checkmark & 0.500$\pm$0.085 & 0.625$\pm$0.058 & 0.517$\pm$0.073 & 0.549$\pm$0.054 & 0.548                          \\
w/o $\mathcal{L}_{\text{sim}}$ & $\ast$ & \checkmark & {\color{red}0.580$\pm$0.123} & 0.634$\pm$0.048 & {\color{blue}{0.553$\pm$0.127}} & {\color{blue}{0.565$\pm$0.058}} & {\color{blue}{0.583}}                          \\
w/o $\mathcal{L}_{\text{align}}$ & $\ast$ & \checkmark & 0.554$\pm$0.079 & {\color{blue}{0.644$\pm$0.044}} & {\color{red}{0.558$\pm$0.134}} & 0.556$\pm$0.070 & 0.578                          \\
\cellcolor[HTML]{EFEFEF}ProSurv & \cellcolor[HTML]{EFEFEF}$\ast$ & \cellcolor[HTML]{EFEFEF}\checkmark & \cellcolor[HTML]{EFEFEF}{\color{blue}{0.578$\pm$0.110}} & \cellcolor[HTML]{EFEFEF}{\color{red}{0.646$\pm$0.046}} & \cellcolor[HTML]{EFEFEF}{0.540$\pm$0.079} &\cellcolor[HTML]{EFEFEF}{\color{red}{0.595$\pm$0.056}} & \cellcolor[HTML]{EFEFEF}{\color{red}0.590}                       \\ \midrule \midrule
w/o Prototypes & \checkmark & \checkmark & 0.648$\pm$0.060 & 0.637$\pm$0.064 & 0.495$\pm$0.059 & {\color{blue}0.623$\pm$0.108} & 0.601                          \\
w/o $\mathcal{L}_{\text{sim}}$ & \checkmark & \checkmark & 0.610$\pm$0.087 & 0.648$\pm$0.047 & {\color{blue}0.531$\pm$0.112} & 0.610$\pm$0.129 & 0.600                          \\
w/o $\mathcal{L}_{\text{align}}$ & \checkmark & \checkmark & {\color{blue}{0.662$\pm$0.058}} & {\color{blue}{0.654$\pm$0.047}} & 0.528$\pm$0.117 & 0.606$\pm$0.124 & {\color{blue}{0.613}}                          \\
\cellcolor[HTML]{EFEFEF}ProSurv & \cellcolor[HTML]{EFEFEF}\checkmark & \cellcolor[HTML]{EFEFEF}\checkmark & \cellcolor[HTML]{EFEFEF}{\color{red}{0.675$\pm$0.070}} & \cellcolor[HTML]{EFEFEF}{\color{red}{0.655$\pm$0.047}} & \cellcolor[HTML]{EFEFEF}{\color{red}{0.609$\pm$0.073}} & \cellcolor[HTML]{EFEFEF}{\color{red}{0.641$\pm$0.088}} & \cellcolor[HTML]{EFEFEF}{\color{red}{0.645}}                          \\ \bottomrule
\end{tabular}}
\label{tab:ablation}
\vspace{-0.3cm}
\end{table}

\subsection{Model Performance Comparison} 
We compared the ProSurv with 11 state-of-the-art survival analysis methods, ranging from image-based \cite{ilse2018attention,li2024dynamic,shao2021transmil,yang2024mambamil}, genome-based  \cite{klambauer2017self}, multimodal \cite{chen2021multimodal,jaume2024modeling,xu2023multimodal,zhou2023cross}, and multimodal training unimodal testing method \cite{wang2025histo} under various scenarios. 

\tabref{tab:quantitative} illustrates the quantitative results, from which we can derive the following observations: (1) Multimodal methods generally outperform unimodal methods, highlighting that multimodal data contains more survival-related information than unimodal data. (2) ProSurv achieves state-of-the-art performance with multimodal inference, obtaining an overall C-index of 0.645. This indicates the superiority of the knowledge enhancement brought by prototype-guided cross-translation. (3) ProSurv excels in inference using unimodal data. When using only pathology images, ProSurv achieves a promising C-index of 0.640, surpassing G-HANet (0.617) and MCAT (0.614) significantly. This demonstrates that the prototypes capture the multimodal knowledge and can effectively enhance the patients' representations for survival prediction using unimodal data.

\begin{figure}[!t]
\centerline{\includegraphics[width=0.95\columnwidth]
{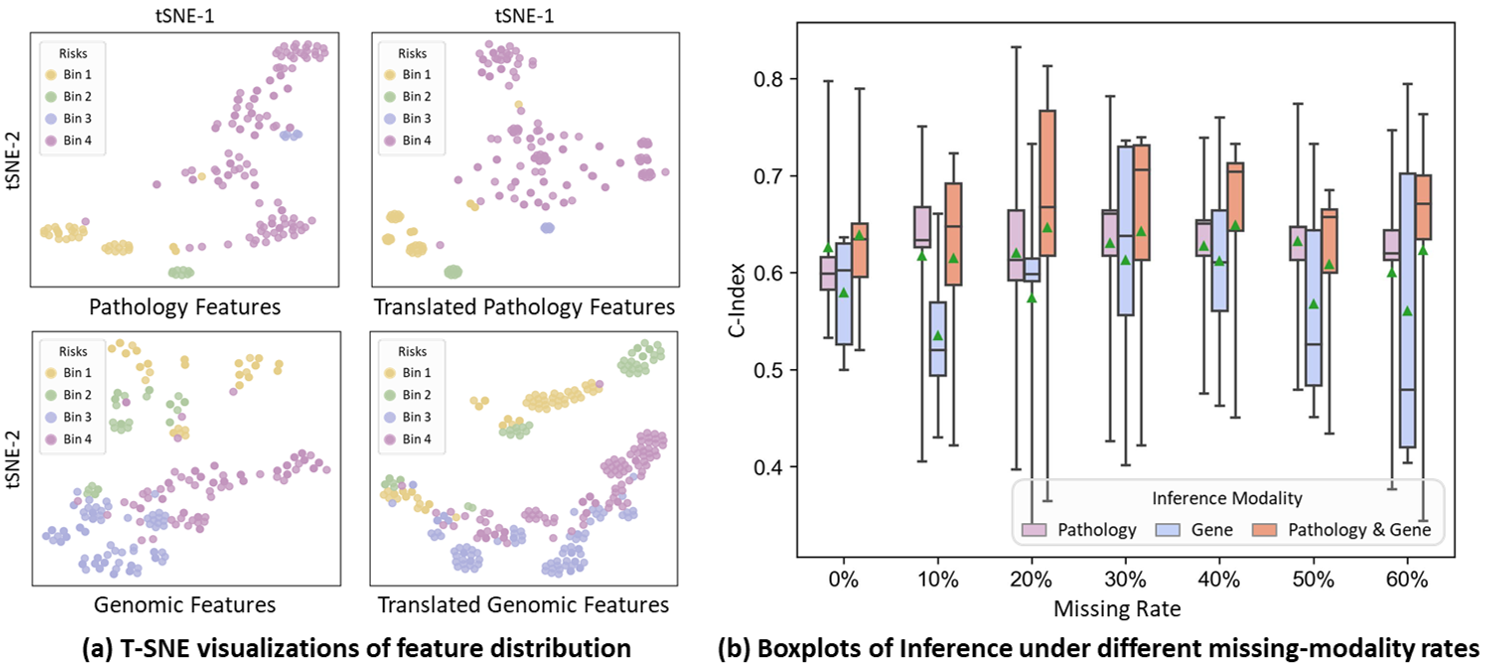}}
\caption{Distribution visualizations of original and translated features (a). Robustness evaluation using a mixture of uni- and multimodal training data on TCGA-CRAD (b).}
\label{fig:ablation}
\vspace{-0.3cm}
\end{figure}

\subsection{Ablation Study}
In this subsection, we perform ablation experiments to validate the effectiveness of each module, with experimental results in \tabref{tab:ablation} and \figref{fig:ablation}. 

\noindent \textbf{Effectiveness of Prototypes.} We replace the cross-modal translation modules with multi-layer perceptrons. The overall C-index decreases by 4.2\% under various scenarios. This is expected, as discarding prototypes means losing modality-specific knowledge, reducing the knowledge throughout the computation process.  

\noindent \textbf{Effectiveness of Intra-Modal Prototype Update.} The use of intra-modal prototype update makes the prototypes effectively capture risk-relevant knowledge, with 3.8\%, 0.7\%, and 4.5\% C-index improvements under pathology-only, gene-only, and complete modality scenes, respectively.

\noindent \textbf{Effectiveness of Prototype-Guided Cross-Modal Translation.} 
Alignment loss $\mathcal{L}_{\text{align}}$ is used to optimize translated features in ProSurv. As shown in \tabref{tab:ablation}, abandoning the loss leads to inferior model performance. Additionally, \figref{fig:ablation}(a) visualizes the feature distributions of original and translated features using t-SNE \cite{van2008visualizing}. The highly identical distributions of the counterpart features reveal the reality of the translated features and the effectiveness of the module. 


\noindent \textbf{Model Robustness Evaluation.} 
Different from existing methods, ProSurv can adaptively learn knowledge from incomplete data. \figref{fig:ablation}(b) illustrates the model performance across varying rates of unimodal data while training the neural network. When the rate is lower than 50\%, the performance of ProSurv does not degrade significantly, showing its powerful learning capability and providing an insightful solution for adaptive survival analysis.

\section{Conclusion}
In this paper, we present ProSurv, a prototype-guided cross-modal knowledge enhancement framework for adaptive survival analysis. The key innovations of ProSurv lie in using prototype banks to generate cross-modal features and achieve an adaptive survival prediction using arbitrary input data. Extensive experiments including performance comparison and ablation study underscore the superiority and robustness of ProSurv. ProSurv addresses a critical practical challenge in computational pathology, offering substantial significance for precision medicine.


%
%
%
%

\bibliographystyle{splncs04}
\bibliography{ref}

\end{document}